\documentclass[10pt,twocolumn,letterpaper]{article}

\usepackage{iccv}
\usepackage{times}
\usepackage{epsfig}
\usepackage{graphicx}
\usepackage{amsmath}
\usepackage{amssymb}
\usepackage{times}
\usepackage{epsfig}
\usepackage{graphicx}
\usepackage{amsmath}
\usepackage{amssymb}

\usepackage{booktabs}
\usepackage{tabulary}
\usepackage[dvipsnames]{xcolor}
\usepackage[caption=false]{subfig}
\usepackage[british,english,american]{babel}
\usepackage{soul}
\usepackage{xspace}\xspace
\usepackage{adjustbox}
\usepackage{array}

\usepackage{dblfloatfix}
\usepackage{transparent}

\usepackage{algorithm}
\usepackage{listings}
\usepackage{multirow}
\newcommand{\red}[1]{\textcolor{red}{#1}}
\newcommand{\gray}[1]{\textcolor{gray}{#1}}
\newcommand{\green}[1]{\textcolor[RGB]{96,177,87}{#1}}
\newcommand{\fn}[1]{\footnotesize{#1}}

\newcommand{\gbf}[1]{\green{\bf{\fn{(#1)}}}}
\newcommand{\rbf}[1]{\gray{\bf{\fn{(#1)}}}}

\usepackage{etoolbox}
\makeatletter
\AfterEndEnvironment{algorithm}{\let\@algcomment\relax}
\AtEndEnvironment{algorithm}{\kern2pt\hrule\relax\vskip3pt\@algcomment}
\let\@algcomment\relax
\newcommand\algcomment[1]{\def\@algcomment{\footnotesize#1}}
\renewcommand\fs@ruled{\def\@fs@cfont{\bfseries}\let\@fs@capt\floatc@ruled
  \def\@fs@pre{\hrule height.8pt depth0pt \kern2pt}%
  \def\@fs@post{}%
  \def\@fs@mid{\kern2pt\hrule\kern2pt}%
  \let\@fs@iftopcapt\iftrue}
\makeatother

\newcommand{\app}{\raise.17ex\hbox{$\scriptstyle\sim$}}

\newcommand{\calT}{\mathcal{T}}
\newcommand{\bI}{\mathbf{I}}
\newcommand{\bP}{\mathbf{P}}

\newcolumntype{x}[1]{>{\centering\arraybackslash}p{#1pt}}
\newcolumntype{y}[1]{>{\raggedright\arraybackslash}p{#1pt}}
\newcolumntype{z}[1]{>{\raggedleft\arraybackslash}p{#1pt}}
\newlength\savewidth\newcommand\shline{\noalign{\global\savewidth\arrayrulewidth
  \global\arrayrulewidth 1pt}\hline\noalign{\global\arrayrulewidth\savewidth}}

\makeatletter\renewcommand\paragraph{\@startsection{paragraph}{4}{\z@}
  {.5em \@plus1ex \@minus.2ex}{-.5em}{\normalfont\normalsize\bfseries}}\makeatother

\usepackage{textpos}

\usepackage[pagebackref=true,breaklinks=true,letterpaper=true,colorlinks,bookmarks=false]{hyperref}

\iccvfinalcopy 

\def\iccvPaperID{6486} 

\ificcvfinal\fi

\begin{document}

\title{DetCo: Unsupervised Contrastive  Learning for Object Detection}

\author{
    Enze Xie$^{1*}$ ,
    Jian Ding$^{3}$\thanks{equal contribution}, 
    Wenhai Wang$^{4}$, 
    Xiaohang Zhan$^{5}$, \\
    Hang Xu$^2$,
    Peize Sun$^1$,
    Zhenguo Li$^2$, 
    Ping Luo$^1$ \\
    $^1$The University of Hong Kong~~~
    $^2$Huawei Noah's Ark Lab\\
    $^3$Wuhan University~~~
    $^4$Nanjing University~~~
    $^5$Chinese University of Hong Kong
}

\maketitle
\ificcvfinal\thispagestyle{empty}\fi

\begin{abstract}
We present DetCo, a simple yet effective self-supervised approach for object detection. Unsupervised pre-training methods have been recently designed for object detection, but they are usually deficient in image classification, or the opposite. 
Unlike them, DetCo transfers well on downstream instance-level dense prediction tasks, while maintaining competitive image-level classification accuracy. The advantages are derived from (1) multi-level supervision to intermediate representations, (2) contrastive learning between global image and local patches. These two designs facilitate discriminative and consistent global and local representation at each level of feature pyramid, improving detection and classification, simultaneously.

Extensive experiments on VOC, COCO, Cityscapes, and ImageNet demonstrate that DetCo not only  outperforms  recent methods on a series of 2D and 3D instance-level detection tasks, but also competitive on image classification.  For example, on ImageNet classification, DetCo is 6.9\% and 5.0\% top-1 accuracy better than  InsLoc and  DenseCL,  which are two contemporary works designed for object detection.  Moreover, on COCO detection, DetCo is 6.9 AP better than SwAV with Mask R-CNN C4. 
Notably, DetCo largely boosts up Sparse R-CNN, a recent strong detector, from 45.0 AP to 46.5 AP (+1.5 AP), establishing a new SOTA on COCO. Code is available.
\end{abstract}

\section{Introduction}
Self-supervised learning of visual representation is an essential problem in computer vision, facilitating many downstream tasks such as image classification, object detection, and semantic segmentation~\cite{resnet,fasterrcnn,pspnet}. 
It aims to provide models pre-trained on large-scale unlabeled data for downstream tasks.
Previous methods focus on designing different pretext tasks.
One of the most promising directions among them is contrastive learning~\cite{oord2018representation}, which transforms one image into multiple views, minimizes the distance between views from the same image, and maximizes the distance between views from different images in a feature map.

\begin{figure}[t]
\begin{center}
\includegraphics[width=0.44\textwidth]{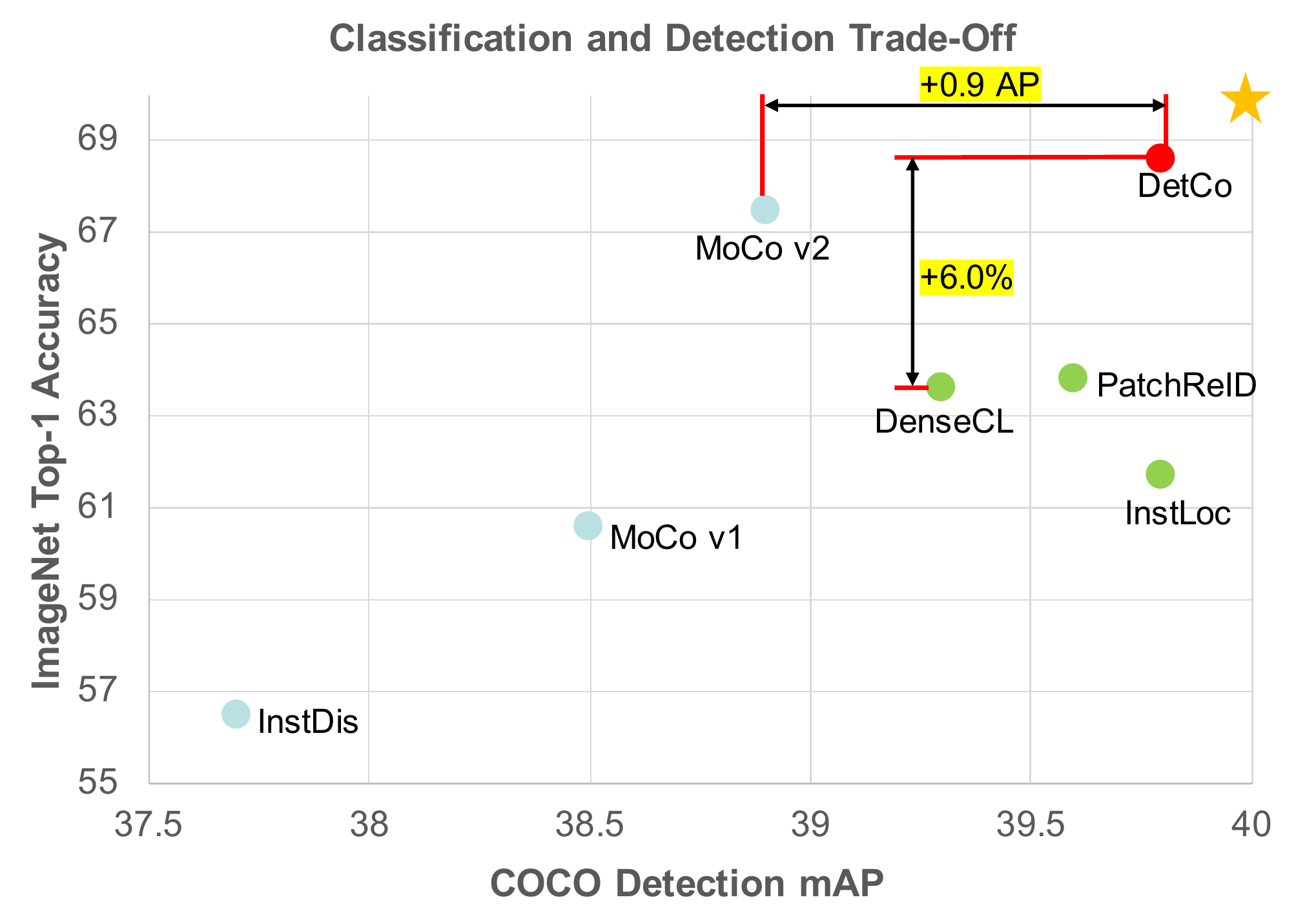}
\caption{\textbf{Transfer accuracy on Classification and Detection.} DetCo achieves the best performance trade-off on both classification and detection. For example, DetCo outperforms its strong baseline, MoCo v2~\cite{mocov2}, by 0.9 AP on COCO detection. 
Moreover, DetCo is significant better than recent work \eg DenseCL~\cite{densecl}, InsLoc~\cite{instloc}, PatchReID~\cite{dupr} on ImageNet classification while also has advantages on object detection. Note that these three methods are concurrent work and specially designed for object detection~(mark with \green{\textbf{green}}). The yellow asterisk indicates that a desired method should have both high performance in detection and classification.
}
\label{fig:trade}
\end{center}
\vspace{-5mm}
\end{figure}

In the past two years, some methods based on contrastive learning and online clustering, \eg MoCo v1/v2 \cite{moco,mocov2}, BYOL \cite{byol}, and SwAV \cite{swav}, have achieved great progress to bridge the performance gap between unsupervised and fully-supervised methods for image classification. However, their transferring ability on object detection is not satisfactory.
Concurrent to our work, recently DenseCL~\cite{densecl}, InsLoc~\cite{instloc} and PatchReID~\cite{dupr} also adopt contrastive learning to design detection-friendly pretext tasks. Nonetheless, these methods only transfer well on object detection but sacrifice image classification performance, as shown in Figure~\ref{fig:trade} and Table~\ref{tab:trade}. 
So,  \textit{it is challenging to design a pretext task that can reconcile instance-level detection and image classification.}

We hypothesize that there is no unbridgeable gap between image-level classification and instance-level detection. Intuitively, image classification recognizes global instance from a single high-level feature map, while object detection recognizes local instance from multi-level feature pyramids. From this perspective, it is desirable to build instance representation that are (1) discriminative at each level of feature pyramid (2) consistent for both global image and local patch ~(\textit{a.k.a} sliding windows). However, existing unsupervised methods overlook these two aspects. Therefore, detection and classification cannot mutually improve.

In this work, we present DetCo, which is a contrastive learning framework beneficial for instance-level detection tasks while maintaining competitive image classification transfer accuracy.
DetCo contains 
(1) multi-level supervision on features from different stages of the backbone network.
(2) contrastive learning between global image and local patches. 
Specifically, the multi-level supervision directly optimizes the features from each stage of backbone network, ensuring strong discrimination in each level of pyramid features. This supervision leads to better performance for dense object detectors by multi-scale prediction.
The global and local contrastive learning guides the network to learn consistent representation on both image-level and patch-level, which can not only keep each local patch highly discriminative but also promote the whole image representation, benefiting both object detection and image classification. 

\begin{table}[]
\hspace{-2mm}
\scalebox{0.75}{
\begin{tabular}{l|c|cc|c|c}
\multicolumn{1}{l|}{\multirow{2}{*}{Method}} & \multirow{2}{*}{Place} & \multicolumn{2}{c|}{ImageNet Cls.}                      & COCO Det. & Cityscapes Seg. \\ \cline{3-6} 
\multicolumn{1}{c|}{}                        &                        & \multicolumn{1}{l}{Top-1} & \multicolumn{1}{l|}{Top-5} & mAP        & mIoU            \\ \shline
MoCo v1\cite{moco}    & CVPR'20 & 60.6 & -    & 38.5 & 75.3 \\ 
MoCo v2\cite{mocov2}    & Arxiv   & 67.5 & -    & 38.9 & 75.7 \\ 
InstLoc\cite{instloc}    & CVPR'21 & 61.7 & -    & 39.8 & -    \\ 
DenseCL\cite{densecl}    & CVPR'21 & 63.6 & 85.8 & 39.3 & 75.7 \\ 
PatchReID\cite{dupr} & Arxiv   & 63.8 & 85.6    & 39.6 & \textbf{76.6} \\ \hline
DetCo      & -       & \textbf{68.6} & \textbf{88.5} & \textbf{39.8} & 76.5 \\ 
\end{tabular}}
\vspace{2mm}
\caption{\textbf{Classification and Detection trade-off for recent detection-friendly self-supervised methods.} 
Compared with concurrent InstLoc\cite{instloc}, DenseCL\cite{densecl} and PatchReID\cite{dupr}, DetCo is significantly better by 6.9\%, 5.0\% and 4.8\% on ImageNet classification. Moreover, DetCo is also on par with these methods on dense prediction tasks, achieving best trade-off.
}
\label{tab:trade}
\vspace{-5mm}
\end{table}

DetCo achieves state-of-the-art transfer performance on various 2D and 3D instance-level detection tasks \eg VOC and COCO object detection, semantic segmentation and DensePose.
Moreover, the performance of DetCo on ImageNet classification and VOC SVM classification is still very competitive. 
For example, as shown in Figure~\ref{fig:trade} and Table~\ref{tab:trade}, 
DetCo improves MoCo v2 on both classification and dense prediction tasks. 
DetCo is significant better than DenseCL~\cite{densecl}, InsLoc~\cite{instloc} and PatchReID~\cite{dupr} on ImageNet classification by 6.9\%, 5.0\% and 4.8\% and slightly better on object detection and semantic segmentation. Please note DenseCL, InsLoc and PatchReID are three concurrent works which are designed for object detection but sacrifice classification. 
Moreover, DetCo boosts up Sparse R-CNN~\cite{sparsercnn}, which is a recent end-to-end object detector without q, from a very high baseline 45.0 AP to 46.5 AP~(+1.5 AP) on COCO dataset with ResNet-50 backbone, establishing a new state-of-the-art detection result. 
In the 3D task, DetCo outperforms ImageNet supervised methods and MoCo v2 in all metrics on COCO DensePose, especially +1.4 on AP$_{50}$.

Overall, the main \textbf{contributions} of this work are three-fold: 

\begin{itemize}
\itemsep -0.1cm
\item 
We introduce a simple yet effective self-supervised pretext task, named DetCo, which is beneficial for instance-level detection tasks. DetCo can utilize large-scale unlabeled data and provide a strong pre-trained model for various downstream tasks.

\item 
Benefiting from the design of multi-level supervision and contrastive learning between global images and local patches, DetCo successfully improves the transferring ability on object detection without sacrificing image classification, compared to contemporary self-supervised counterparts. 

\item 
Extensive experiments on PASCAL VOC \cite{voc}, COCO \cite{coco} and Cityscapes \cite{cityscape} show that DetCo outperforms previous state-of-the-art methods when transferred to a series of 2D and 3D instance-level detection tasks, \eg object detection, instance segmentation, human pose estimation, DensePose, as well as semantic segmentation. 
\end{itemize}

\section{Related Work}
Existing unsupervised methods for representation learning can be roughly divided into two classes, generative and discriminative.
Generative methods~\cite{donahue2016adversarial,dumoulin2016adversarially,donahue2019large,brock2018large} typically rely on auto-encoding of images~\cite{vincent2008extracting,kingma2013auto,rezende2014stochastic} or adversarial learning~\cite{goodfellow2014generative}, and operate directly in pixel space.
Therefore, most of them are computationally expensive, and the pixel-level details required for image generation may not be necessary for learning high-level representations.

\begin{figure*}[!t]
\begin{center}
\scalebox{0.9}{
\includegraphics[width=0.9\textwidth]{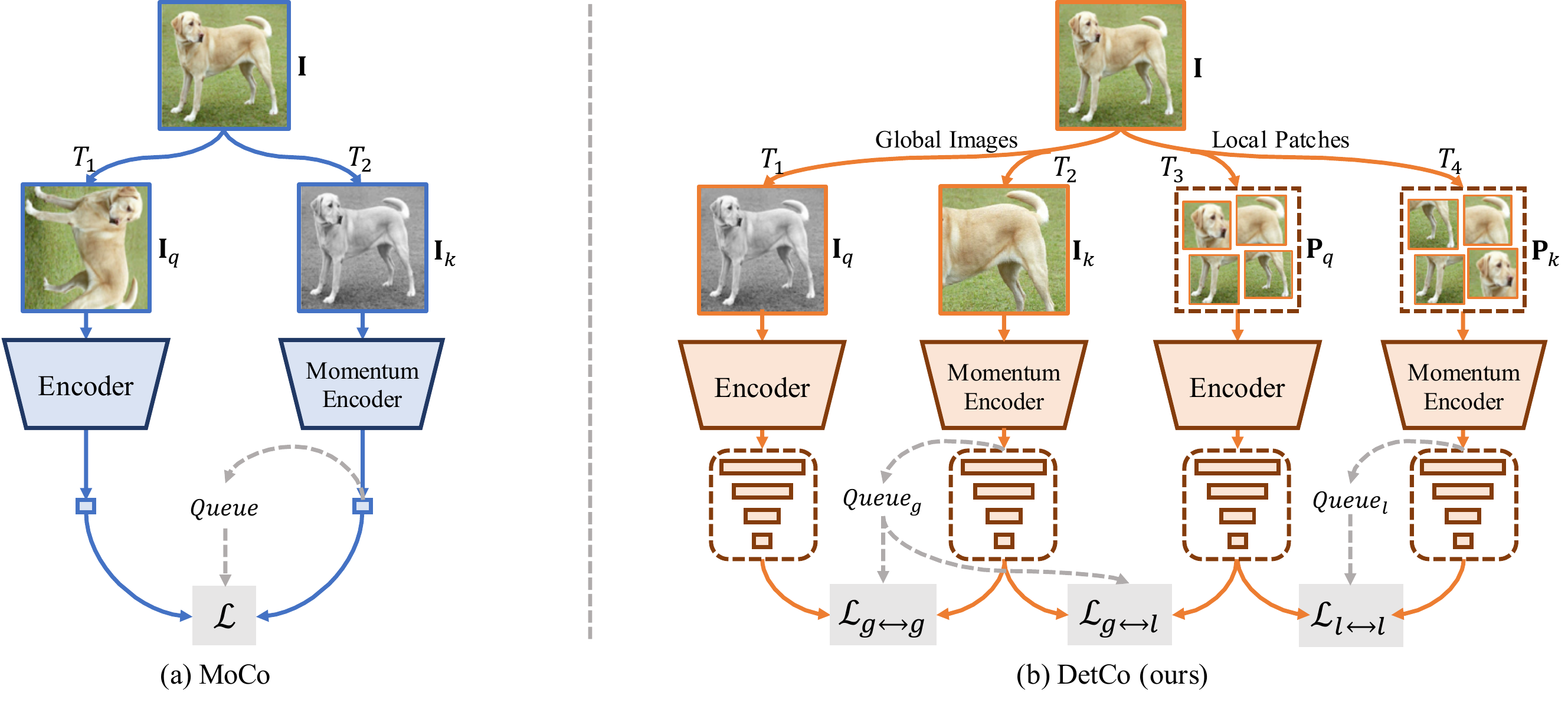}}
\caption{
\textbf{The overall pipeline of DetCo compared with MoCo~\cite{moco}.} 
(a) is MoCo's framework, which only considers the single high-level feature and learning contrast from a global perspective. 
(b) is our DetCo, which learns representation with multi-level supervision and adds two additional local patch sets for input, building contrastive loss cross the global and local views. 
Note that ``$T$'' means image transforms. 
``$Queue_{g/l}$'' means different memory banks~\cite{wu2018unsupervised} for global/local features.}
\label{fig:pipeline}
\end{center}
\end{figure*}

Among discriminative methods~\cite{relloc,mocov2}, self-supervised contrastive learning~\cite{mocov2,moco,mocov2,swav,byol} currently achieved state-of-the-art performance, arousing extensive attention from researchers.
Unlike generative methods, contrastive learning avoids the computation-consuming generation step by pulling representations of different views of the same image (\emph{i.e.}, positive pairs) close, and pushing representations of views from different images (\emph{i.e.}, negative pairs) apart.
Chen \emph{et al}.~\cite{mocov2} developed a simple framework, termed SimCLR,
for contrastive learning of visual representations.
It learns features by contrasting images after a composition of data augmentations.
After that, He \emph{et al}.~\cite{moco} and Chen \emph{et al}.~\cite{mocov2} proposed MoCo and MoCo v2, using a moving average network (momentum encoder) to maintain consistent representations of negative pairs drawn from a memory bank.
Recently, SwAV~\cite{swav} introduced online clustering into contrastive learning, without requiring to compute pairwise comparisons.
BYOL~\cite{byol} avoided the use of negative pairs by bootstrapping the outputs of a network iteratively to serve as targets for an enhanced representation.

Moreover, earlier methods rely on all sorts of pretext tasks to learn visual representations.
Relative patch prediction~\cite{relloc,doersch2017multi}, colorizing gray-scale images~\cite{zhang2016colorful,larsson2016learning}, image inpainting~\cite{pathak2016context}, image jigsaw puzzle~\cite{noroozi2016unsupervised}, image super-resolution~\cite{ledig2017photo}, and geometric transformations~\cite{dosovitskiy2014discriminative,gidaris2018unsupervised} have been proved to be useful for representation learning.

Nonetheless, most of the aforementioned methods are specifically designed for image classification while neglecting object detection. 
Concurrent to our work, recently DenseCL~\cite{densecl}, InsLoc~\cite{instloc} and PatchReID~\cite{dupr} design pretext tasks for object detection.
However, their transferring performance is poor on image classification.
Our work focuses on designing a better pretext task which is not only beneficial for instance-level detection, but also maintains strong representation for image classification.

\section{Methods}

In this section, we first briefly introduce the overall architecture of the proposed DetCo showed in Figure~\ref{fig:pipeline}.
Then, we present the design of multi-level supervision that keeps features at multiple stages discriminative.
Next, we introduce global and local contrastive learning 
to enhance global and local representation. 
Finally, we provide the implementation details of DetCo.

\subsection{DetCo Framework}\label{sec:detco}
DetCo is a simple pipeline designed mainly based on a strong baseline MoCo v2. 
It composes of a backbone network, a series of MLP heads and memory banks. 
The setting of MLP head and memory banks are same as MoCo v2 for simplicity.
The overall architecture of DetCo is illustrated in Figure~\ref{fig:pipeline}. 

Specifically, DetCo has two simple and effective designs which are different from MoCo v2. 
(1) multi-level supervision to keep features at multiple stages discriminative.
(2) global and local contrastive learning to enhance both global and local feature representation.
The above two different designs make DetCo not only successfully inherit  the advantages of MoCo v2 on image classification but also transferring much stronger on instance-level detection tasks.

The complete loss function of DetCo can be defined as follows:
\begin{equation}
{\mathcal{L}(\bI_q,\bI_k,\bP_q,\bP_k) = \sum_{i=1}^{4} w_i {\cdot} (
\mathcal{L}_{g \leftrightarrow g}^i + 
\mathcal{L}_{l \leftrightarrow l}^i + 
\mathcal{L}_{g \leftrightarrow l}^i
}),
\label{eq:total}
\end{equation}
where $\bI$ represents a global image and $\bP$ represents the local patch set.
Eqn.~\ref{eq:total} is a multi-stage contrastive loss.
In each stage, there are three cross local and global contrastive losses.
We will describe the multi-level supervision $\sum_{i=1}^{4} w_i {\cdot} \mathcal{L}^i$ in Section~\ref{sec:multiloss}, the global  and local contrastive learning $\mathcal{L}_{g \leftrightarrow g}^i+\mathcal{L}_{l \leftrightarrow l}^i+\mathcal{L}_{g \leftrightarrow l}^i$ in Section~\ref{sec:crossloss}.

\subsection{Multi-level Supervision}\label{sec:multiloss}
Modern object detectors predict objects in different levels, \eg RetinaNet and Faster R-CNN FPN. 
They require the features at each level to keep strong discrimination.
To meet the above requirement, we make a simple yet effective modification to the original MoCo baseline.

Specifically, we feed one image to a standard backbone ResNet-50, and it outputs features from different stages, 
termed {\tt Res2, Res3, Res4, Res5}. 
Unlike MoCo that only uses {\tt Res5}, we utilize all levels of features to calculate contrastive losses, ensuring that each stage of the backbone produces discriminative representations.

Given an image $\bI \in \mathbb{R}^{H \times W \times 3}$, it is first transformed to two views of the image $\bI_q$ and $\bI_k$ with 
two transformations randomly drawn from a set of transformations on global views, termed $\calT_g$.
We aim at training an $\tt encoder_q$ together with an $\tt encoder_k$ with the same architecture, where $\tt encoder_k$ update weights using a momentum update strategy~\cite{moco}.
The $\tt encoder_q$ contains a backbone and four global MLP heads to extract features from four levels.
We feed $\bI_q$ to the backbone $b_{q}^\theta(\cdot)$, with parameters $\theta$ that extracts features 
$\{f_{2}, f_{3}, f_{4}, f_{5}\} = b_{q}^\theta(\bI_q)$, where $f_i$ means the feature from the i-th stage.
After obtaining the multi-level features, we append four global MLP heads 
$\{mlp_{q}^2(\cdot), mlp_{q}^3(\cdot), mlp_{q}^4(\cdot), mlp_{q}^5(\cdot)\}$
whose weights are non-shared.
As a result, we obtain four global representations
$\{q_2^g, q_3^g, q_4^g, q_5^g\} = \tt encoder_q(\bI_q)$.
Likewise, we can easily get 
$\{k_2^g, k_3^g, k_4^g, k_5^g\} = \tt encoder_k(\bI_k)$.

MoCo uses InfoNCE to calculate contrasitive loss, formulated as:
\begin{equation}
\mathcal{L}_{g \leftrightarrow g}(\bI_q,\bI_k) = -\log \frac{\exp(q^g{\cdot}k^g_+ / \tau)}{\sum_{i=0}^{K}\exp(q^g{\cdot}k_i^g  / \tau)},
\label{eq:g2g}
\end{equation}
where $\tau$ is a temperature hyper-parameter \cite{wu2018unsupervised}.
We extend it to multi-level contrastive losses for multi-stage features, formulated as:
\begin{equation}
{Loss = \sum_{i=1}^{4} w_i {\cdot} \mathcal{L}_{g \leftrightarrow g}^i},
\label{eq:multi}
\end{equation}
where $w$ is the loss weight, and $i$ indicates the current stage.
Inspired by the loss weight setting in PSPNet~\cite{pspnet},
we set the loss weight of shallow layers to be smaller than deep layers.
In addition, we build an individual memory bank $queue_i$ for each layer.
In the appendix, we provide the pseudo-code of intermediate contrastive loss.

\subsection{Global and Local Contrastive Learning}\label{sec:crossloss}

Modern object detectors repurpose classifiers on local regions~(\textit{a.k.a} sliding windows) to perform detection. So, it requires each local region to be discriminative for instance classification.
To meet the above requirement, we develop global and local contrastive learning to keep consistent instance representation on both patch set and the whole image. This strategy takes advantage of image-level representation to enhance instance-level representation, vice versa.

In detail, we first transform the input image into 9 local patches using jigsaw augmentation, the augmentation details are shown in section~\ref{sec:detail}.
These patches pass through the encoder, and then we can get 9 local feature representation.
After that, we combine these features into one feature representation by a MLP head, and build a cross global-and-local contrastive learning.

Given an image $\bI \in \mathbb{R}^{H \times W \times 3}$, first it is transformed into two local patch set $\bP_q$ and $\bP_k$ by two transformations selected from a local transformation set, termed $\calT_l$.
There are 9 patches $\{p_1, p_2, ..., p_9\}$ in each local patch set.
We feed the local patch set to backbone and  get 9 features 
$F_p = \{f_{p1}, f_{p2}, ..., f_{p9}\}$
at each stage.
Taking a stage as an example,
we build a MLP head for local patch, denoted as $mlp_{local}(\cdot)$, which does not share weights with $mlp_{global}(\cdot)$ in section~\ref{sec:multiloss}. 
Then, $F_p$ is concatenated and fed to the local patch MLP head to get final representation $q^l$. 
Likewise, we can use the same approach to get $k^l$.

The contrastive cross loss has two parts: the global$\leftrightarrow$local contrastive loss and the local$\leftrightarrow$local contrastive loss. The global$\leftrightarrow$local contrastive loss can be written as:
\begin{equation}
\mathcal{L}_{g \leftrightarrow l}(\bP_q,\bI_k) = -\log \frac{\exp(q^l{\cdot}k^g_+ / \tau)}{\sum_{i=0}^{K}\exp(q^l{\cdot}k_i^g  / \tau)}.
\label{eq:g2l}
\end{equation}
Similarly,
the local$\leftrightarrow$local contrastive loss can be formulated as:
\begin{equation}
\mathcal{L}_{l \leftrightarrow l}(\bP_q, \bP_k) = -\log \frac{\exp(q^l{\cdot}k^l_+ / \tau)}{\sum_{i=0}^{K}\exp(q^l{\cdot}k_i^l  / \tau)}.
\label{eq:l2l}
\end{equation}
By learning representations between global image and local patches, 
the instance discrimination of image-level and instance-level are mutually improved.
As a result, both the detection and classification performance boost up.

\subsection{Implementation Details}\label{sec:detail}
We use OpenSelfSup~\footnote{https://github.com/open-mmlab/OpenSelfSup} as the codebase. 
We use a batch size of 256 with 8 Tesla V100 GPUs per experiment. %
We follow the most hyper-parameters settings of MoCo v2.
For data augmentation, the global view augmentation is almost the same as MoCo v2~\cite{mocov2} with random crop and resized to $224\times224$ with a random horizontal flip, gaussian blur and color jittering related to brightness, contrast, saturation, hue and grayscale. Rand-Augmentation\cite{randaugment} is also used on global view. The local patch augmentation follows PIRL~\cite{pirl}. First, a random region is cropped with at least 60\% of the image and resized to $255\times255$, followed by random flip, color jitter and blur, sharing the same parameters with global augmentation. Then we divide the image into $3\times3$ grids and randomly shuffle them; each grid is $85\times85$. A random crop is applied on each patch to get $64\times64$ to avoid continuity between patches. Finally, we obtain nine randomly shuffled patches. 
For a fair comparison, we use standard ResNet-50~\cite{resnet} for all experiments. 
Unless other specified, we pre-train 200 epochs on ImageNet for a fair comparison.

\section{Experiments}\label{sec:exp}
\label{sec:exp}

\noindent We evaluate DetCo on a series of 2D and 3D dense prediction tasks, \eg PASCAL VOC detection, COCO detection, instance segmentation, 2D pose estimation, DensePose and Cityscapes instance and semantic segmentation. We see that DetCo  outperforms existing self-supervised and supervised methods. 

\subsection{Object Detection}

\noindent \textbf{Experimental Setup}. 
We choose three representative detectors: Faster R-CNN~\cite{fasterrcnn}, Mask R-CNN~\cite{maskrcnn} RetinaNet~\cite{focalloss}, and a recent strong detector: Sparse R-CNN~\cite{sparsercnn}. 
Mask R-CNN is two-stage and RetinaNet is one stage detector. Sparse R-CNN is an end-to-end detector without NMS, and it is also state-of-the-art with high mAP on COCO.
Our training settings are the same as MoCo~\cite{moco} for a fair comparison, including using ``SyncBN''~\cite{megdet} in backbone and FPN. 

\begin{table*}[]
\centering
\scalebox{0.75}{
\begin{tabular}{l|l l l |l l l |l l l |l l l}
\multirow{2}{*}{Method} & \multicolumn{6}{c|}{Mask R-CNN R50-C4 COCO 12k} & \multicolumn{6}{c}{Mask R-CNN R50-FPN COCO 12k} \\ \cline{2-13} 
 & AP$^{bb}$ & AP$^{bb}_{50}$ & AP$^{bb}_{75}$ & AP$^{mk}$ & AP$^{mk}_{50}$ & AP$^{mk}_{75}$ & AP$^{bb}$ & AP$^{bb}_{50}$ & AP$^{bb}_{75}$ & AP$^{mk}$ & AP$^{mk}_{50}$ & AP$^{mk}_{75}$ \\ \shline
\gray{Rand Init} &7.9 &16.4 &6.9 &7.6 &14.8 &7.2 &10.7 &20.7 &9.9 &10.3 &19.3 &9.6 \\
Supervised  &27.1 &46.8 &27.6 &24.7 &43.6 &25.3 &28.4 &48.3 &29.5 &26.4 &45.2 &25.7 \\ \hline
InsDis\cite{wu2018unsupervised} &25.8\rbf{-1.3} &43.2\rbf{-3.6} &27.0\rbf{-0.6} &23.7\rbf{-1.0} & 40.4\rbf{-3.2} & 24.5\rbf{-0.8} & 24.2\rbf{-4.2} &41.5\rbf{-6.8} &25.1\rbf{-4.4}&22.8\rbf{-3.6}&38.9\rbf{-6.3}&23.7\rbf{-2.0} \\ 
PIRL\cite{pirl} & 25.5\rbf{-1.6} & 42.6\rbf{-4.2} & 26.8\rbf{-0.8} & 23.2\rbf{-1.5} & 39.9\rbf{-3.7} & 23.9\rbf{-1.4} & 23.7\rbf{-4.7}& 40.4\rbf{-7.9}& 24.4\rbf{-5.1}& 22.1\rbf{-4.3}& 37.9\rbf{-7.3}& 22.7\rbf{-3.0} \\ 
SwAV\cite{swav} &16.5\rbf{-10.6} &35.2\rbf{-11.6} &13.5\rbf{-14.1} &16.1\rbf{-8.6} &32.0\rbf{-11.6} &14.6\rbf{-10.7} &25.5\rbf{-2.9} &46.2\rbf{-2.1} &25.4\rbf{-4.1} &24.8\rbf{-1.6} &43.5\rbf{-1.7} &25.3\rbf{-0.4} \\ 
MoCo\cite{moco} &26.9\rbf{-0.2} &44.5\rbf{-2.3} & 28.2\gbf{+0.6} &24.6\rbf{-0.1} &41.8\rbf{-1.8} &25.6\gbf{+0.3} &25.6\rbf{-2.8} &43.4\rbf{-4.9} &26.6\rbf{-2.9} &23.9\rbf{-2.5} &40.8\rbf{-4.4} &24.8\rbf{-0.9} \\ 
MoCov2\cite{mocov2} &27.6\gbf{+0.5} &45.3\rbf{-1.5} & 28.9\gbf{+1.3} &25.1\gbf{+0.4} &42.6\rbf{-1.0} &26.3\gbf{+1.0} &26.6\rbf{-1.8} &44.9\rbf{-3.4} &27.7\rbf{-1.8} &24.8\rbf{-1.6} &42.1\rbf{-3.1} &25.7\rbf{0.0} \\ \hline
DetCo &\bf{29.8}\gbf{+2.7} &\bf{49.1}\gbf{+2.3} &\bf{31.4}\gbf{+3.8} &\bf{26.9}\gbf{+2.2} &\bf{46.0}\gbf{+2.4} &\bf{27.9}\gbf{+2.6} &\bf{29.6}\gbf{+1.2} &\bf{49.4}\gbf{+1.1} &\bf{31.0}\gbf{+1.5} &\bf{27.6}\gbf{+1.2} &\bf{46.6}\gbf{+1.4} &\bf{28.7}\gbf{+3.0} \\

\end{tabular}}
\caption{\small{
\textbf{Object detection and instance segmentation fine-tuned on COCO}. All methods are pretrained 200 epochs on ImageNet. \textbf{\green{Green}} means increase and \textbf{\gray{gray}} means decrease. DetCo outperforms all supervised and unsupervised counterparts.  
}
}

\label{tab:coco_12k_c4fpn}
\end{table*}
\begin{table*}[]
\centering
\scalebox{0.75}{
\begin{tabular}{l|l l l |l l l |l l l |l l l}
\multirow{2}{*}{Method} & \multicolumn{6}{c|}{Mask R-CNN R50-C4 COCO 90k} & \multicolumn{6}{c}{Mask R-CNN R50-FPN COCO 90k} \\ \cline{2-13} 
 & AP$^{bb}$ & AP$^{bb}_{50}$ & AP$^{bb}_{75}$ & AP$^{mk}$ & AP$^{mk}_{50}$ & AP$^{mk}_{75}$ & AP$^{bb}$ & AP$^{bb}_{50}$ & AP$^{bb}_{75}$ & AP$^{mk}$ & AP$^{mk}_{50}$ & AP$^{mk}_{75}$ \\ \shline
\gray{Rand Init} &26.4 &44.0 &27.8 &29.3 &46.9 &30.8 &31.0 &49.5 &33.2 &28.5 &46.8 &30.4 \\
Supervised  &38.2 &58.2 &41.2 &33.3 &54.7 &35.2 &38.9 &59.6 &42.7 &35.4 &56.5 &38.1  \\ \hline
InsDis\cite{wu2018unsupervised}      & 37.7\rbf{-0.5} & 57.0\rbf{-1.2} & 40.9\rbf{-0.3} & 33.0\rbf{-0.3} & 54.1\rbf{-0.6} & 35.2\rbf{0.0} & 37.4\rbf{-1.5} & 57.6\rbf{-2.0}  & 40.6\rbf{-2.1} &34.1\rbf{-1.3} & 54.6\rbf{-1.9} & 36.4\rbf{-1.7} \\
PIRL\cite{pirl}        & 37.4\rbf{-0.8} & 56.5\rbf{-1.7} & 40.2\rbf{-1.0} & 32.7\rbf{-0.6} & 53.4\rbf{-1.3} & 34.7\rbf{-0.5} & 37.5\rbf{-1.4} & 57.6\rbf{-2.0} & 41.0\rbf{-1.7} &34.0\rbf{-1.4} & 54.6\rbf{-1.9} & 36.2\rbf{-1.9} \\ 
SwAV\cite{swav}        & 32.9\rbf{-5.3} & 54.3\rbf{-3.9} & 34.5\rbf{-6.7} & 29.5\rbf{-3.8} & 50.4\rbf{-4.3} & 30.4\rbf{-4.8} & 38.5\rbf{-0.4} & 60.4\gbf{+0.8} & 41.4\rbf{-1.3} &35.4\rbf{0.0}  & 57.0\gbf{+0.5} & 37.7\rbf{-0.4} \\ 
MoCo\cite{moco}     &38.5\gbf{+0.3} &58.3\gbf{+0.1} &41.6\gbf{+0.4} &33.6\gbf{+0.3} &54.8\gbf{+0.1} &35.6\gbf{+0.4} &38.5\rbf{-0.4} &58.9\rbf{-0.7}         & 42.0\rbf{-0.7} &35.1\rbf{-0.3} & 55.9\rbf{-0.6} &37.7\rbf{-0.4} \\ 
MoCov2\cite{mocov2}     &38.9\gbf{+0.7} &58.4\gbf{+0.2} &42.0\gbf{+0.8} &34.2\gbf{+0.9} &55.2\gbf{+0.5} &36.5\gbf{+1.3} &38.9\rbf{0.0}  &59.4\rbf{-0.2}         &42.4\rbf{-0.3}  &35.5\gbf{+0.1} & 56.5\rbf{0.0} &38.1\rbf{0.0} \\ \hline
DetCo      &\bf{39.8}\gbf{+1.6} &\bf{59.7}\gbf{+1.5} &\bf{43.0}\gbf{+1.8} &\bf{34.7}\gbf{+1.4} &\bf{56.3}\gbf{+1.6} &\bf{36.7}\gbf{+1.5} &\bf{40.1}\gbf{+1.2} &\bf{61.0}\gbf{+1.4}         &\bf{43.9}\gbf{+1.2}  &\bf{36.4}\gbf{+1.0} & \bf{58.0}\gbf{+1.5} &\bf{38.9}\gbf{+0.8} \\ 
\end{tabular}}
\caption{\small{
\textbf{Object detection and instance segmentation fine-tuned on COCO}. 
All methods are pretrained 200 epochs on ImageNet.
DetCo outperforms all supervised and unsupervised counterparts. 
}
}
\label{tab:coco_90k_c4fpn}
\vspace{-3mm}
\end{table*}

\begin{figure}[t]
\begin{center}
\includegraphics[width=0.44\textwidth]{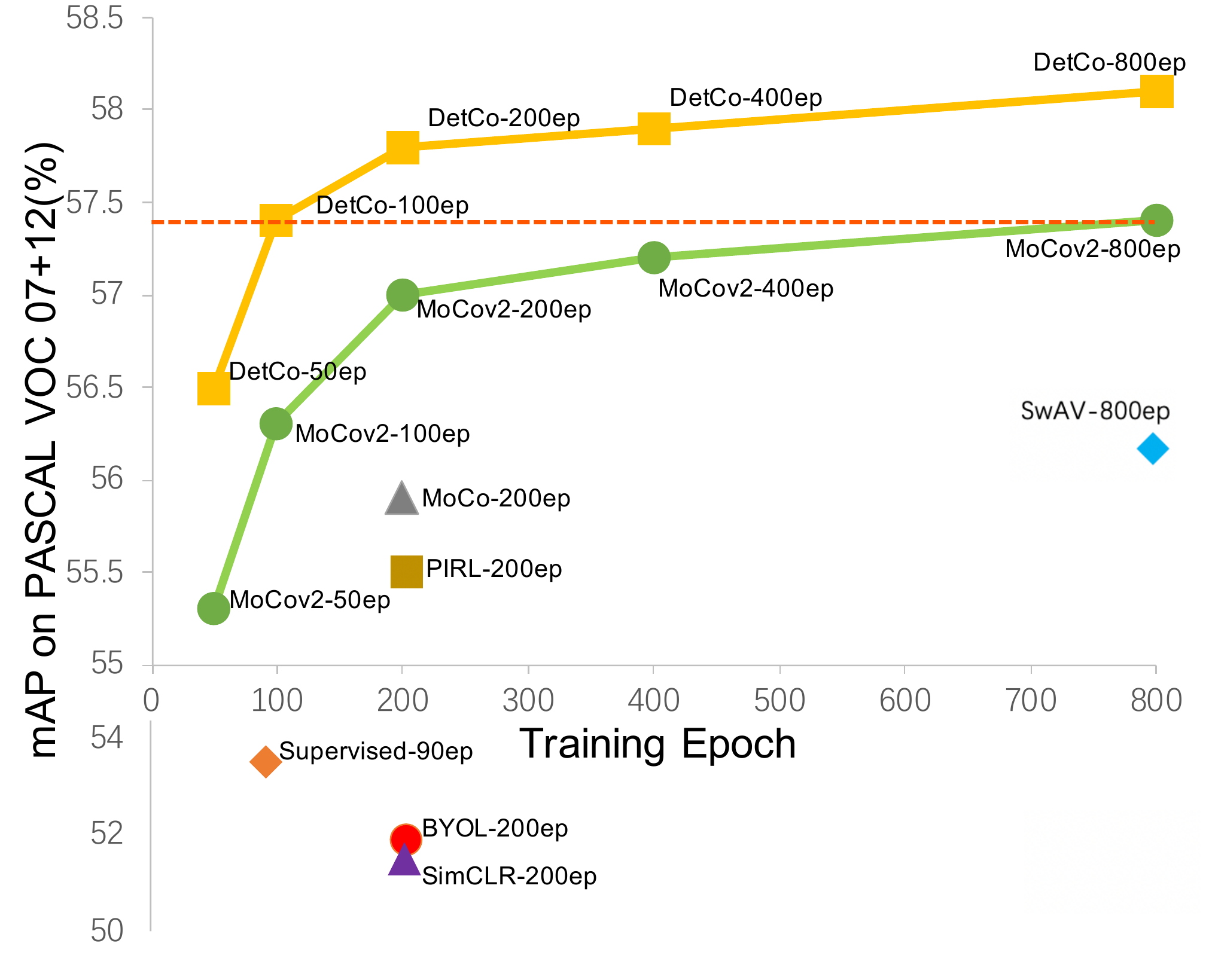}
\caption{\textbf{Comparisons of mAP on PASCAL VOC 07+12 object detection.} For different pre-training epoches, we see that DetCo consistently outperforms MoCo v2\cite{mocov2}, which is a strong competitor on VOC compared to other methods. For example, DetCo-100ep already achieves similar mAP compared to MoCov2-800ep. Moreover, DetCo-800ep achieves state-of-the-art and outperforms other counterparts.}
\label{fig:performance}
\end{center}
\end{figure}

\noindent \textbf{PASCAL VOC.} 
As shown in Table~\ref{tab:voc} and Figure~\ref{fig:performance}, MoCo v2 is a strong baseline, which has already surpassed other unsupervised learning methods in VOC detection. However, our DetCo consistently outperforms the MoCo v2 at 200 epochs and 800 epochs. More importantly, with only 100 epoch pre-training, DetCo achieves almost the same performance as MoCo v2-800ep~(800 epoch pre-training). Finally, DetCo-800ep establishes the new state-of-the-art, 58.2 in mAP and 65.0 in AP$_{75}$, which brings \textbf{4.7} and \textbf{6.2} improvements in AP and AP$_{75}$ respectively, compared with supervised counterpart. The improvements on the more stringent AP$_{75}$ are much larger than the AP, indicating that the intermediate and patch contrasts are beneficial to the localization.

\noindent \textbf{COCO with 1$\times$ and 2$\times$ Schedule.} 
Table~\ref{tab:coco_90k_c4fpn} shows the Mask RCNN~\cite{maskrcnn} results on 1$\times$ schedule, 
DetCo outperforms MoCo v2 baseline by 0.9 and 1.2 AP for R50-C4 and R50-FPN backbones. It also outperforms the supervised counterpart by \textbf{1.6} and \textbf{1.2} AP for R50-C4 and R50-FPN respectively. The results of 2$\times$ schedule is in Appendix.
The column 2-3 of Table~\ref{tab:retina_pose} shows the results of one stage detector RetinaNet. DetCo pretrain is 1.0 and 1.2 AP better than supervised methods and MoCo v2. DetCo is also \textbf{1.3} higher than MoCov2 on AP$_{50}$ with 1$\times$ schedule.

\vspace{2mm}
\noindent \textbf{COCO with Few Training Iterations.} 
COCO is much larger than PASCAL VOC in the data scale. Even training from scratch~\cite{he2019rethinking} can get a satisfactory result. 
To verify the effectiveness of unsupervised pre-training, we conduct experiments on extremely stringent conditions: only train detectors with 12k iterations($\approx$ 1/7$\times$ \textit{vs.} 90k-1$\times$ schedule). 
The 12k iterations make detectors heavily under-trained and far from converge, as shown in Table~\ref{tab:coco_12k_c4fpn} and Table~\ref{tab:retina_pose} column 1. 
Under this setting, for Mask RCNN-C4, DetCo exceeds MoCo v2 by \textbf{3.8} AP in AP$_{50}^{bb}$ and outperforms supervised methods in all metrics, which indicates DetCo can significantly fasten the training convergence. 
For Mask RCNN-FPN and RetinaNet, DetCo also has significant advantages over MoCo v2 and supervised counterpart.

\vspace{2mm}
\noindent \textbf{COCO with Semi-Supervised Learning.} 
Transferring to a small dataset has more practical value. As indicated in the ~\cite{rethinking}, when only use 1\% data of COCO, the train from scratch's performance can not catch up in mAP with ones that have pre-trained initialization. To verify the effectiveness of self-supervised learning on a small-scale dataset, we randomly sample 1\%, 2\%, 5\%, 10\% data to fine-tune the RetinaNet. For all the settings, we fine-tune the detectors with 12k iterations to avoid overfitting. Other settings are the same as COCO 1$\times$ and 2$\times$ schedule. 

The results for RetinaNet with 1\%, 2\%, 5\%, 10\% are shown in Table~\ref{tab:retina_1_2_5_10}.
We find that in four semi-supervised settings, DetCo significantly surpasses the supervised counterpart and MoCo v2 strong baseline. For instance, DetCo outperforms the supervised method by \textbf{2.3} AP and MoCo v2 by \textbf{1.9} AP when using 10\% data. 
These results show that the DetCo pre-trained model is also beneficial for semi-supervised object detection.
More results for Mask R-CNN with 1\%, 2\%, 5\%, and 10\% data are in the appendix.

\vspace{2mm}
\noindent \textbf{DetCo + Recent Advanced Detector.}
In table~\ref{tab:sparsercnn}, we find that DetCo can improve Sparse R-CNN\cite{sparsercnn} with \textbf{1.5} mAP and \textbf{3.1} AP$_s$. Sparse R-CNN is a recent strong end-to-end detector with high performance, and DetCo can further largely boost up Sparse R-CNN's performance and achieved the new state of the arts on COCO with \textbf{46.5} AP. 

\vspace{2mm}
\noindent \textbf{DetCo \vs Concurrent SSL Methods.} 
InsLoc\cite{instloc}, DenseCL\cite{densecl} and PatchReID\cite{dupr} are recent works designed for object detection. They improved the performance of object detection but largely sacrifice the performance of image classification.
As shown in Table~\ref{tab:trade}, DetCo has significant advantages than InsLoc, DenseCL and PatchReID on ImageNet classification by \textbf{+6.9\%}, \textbf{+5.0\%} and \textbf{+4.8\%}. Moreover, on COCO detection, DetCo is also better than these methods. 

\begin{table}[]
\centering
\scalebox{0.9}{
\begin{tabular}{l|l|l|l|l|l|l}
          & AP   & AP$_{50}$ & AP$_{75}$ & AP$_{s}$ & AP$_{m}$  & AP$_{l}$  \\ \shline
Supervised   & 45.0 & 64.1 & 49.0 & 27.7 & 47.5 & 59.6 \\ \hline
DetCo & \bf{46.5} & \bf{65.7} & \bf{50.8} & \bf{30.8} & \bf{49.5} & \bf{59.7} \\ 
\end{tabular}}
\caption{\textbf{DetCo \vs Supervised pre-train} on Sparse R-CNN. DetCo largely improves 1.5 mAP and 3.1 AP$_s$.}
\label{tab:sparsercnn}
\end{table}
\begin{table}[t]
\centering
\scalebox{0.9}{
\begin{tabular}{l|c|l|l|l}
Method       & Epoch & AP$^{dp}$ & AP$^{dp}_{50}$ & AP$^{dp}_{75}$ \\ \shline
\gray{Rand Init} & -          &40.8             &78.6            &37.3       \\ 
Supervised   & 90             &50.8             &86.3            &52.6 \\ \hline
MoCo~\cite{moco}      & 200   &49.6\rbf{-1.2}  &85.9\rbf{-0.4} &50.5\rbf{-2.1}       \\ 
MoCo v2~\cite{mocov2} & 200   &50.9\gbf{+0.1}  &87.2\gbf{+0.9} &52.9\gbf{+0.3}       \\  \hline
DetCo        & 200   &\bf{51.3}\gbf{+0.5}           &\bf{87.7}\gbf{+1.4} &\bf{53.3}\gbf{+0.7} \\ 
\end{tabular}}
\caption{\textbf{DetCo \textit{vs.} other methods on Dense Pose task.} It also performs best on monocular 3D human shape prediction.}
\label{tab:dprcnn}
\end{table}

\begin{table}[t]
\centering
\scalebox{0.93}{
\begin{tabular}{l|l l|l}
\multicolumn{1}{c|}{\multirow{2}{*}{Methods}} & \multicolumn{2}{c|}{Instance Seg.} & \multicolumn{1}{c}{Semantic Seg.} \\ \cline{2-4} 
 & \multicolumn{1}{l}{AP$^{mk}$} & \multicolumn{1}{l|}{AP$_{50}^{mk}$} & \multicolumn{1}{l}{mIOU} \\ \shline
\gray{Rand Init} & 25.4 & 51.1 & 65.3 \\ 
supervised   & 32.9                & 59.6                 & 74.6 \\ \hline
InsDis~\cite{wu2018unsupervised}     & 33.0~\gbf{+0.1} & 60.1~\gbf{+0.5}  & 73.3~\rbf{-1.3} \\ 
PIRL~\cite{pirl}         & 33.9~\gbf{+1.0} & 61.7~\gbf{+2.1}   & 74.6~\rbf{0.0} \\ 
SwAV~\cite{swav}         & 33.9~\gbf{+1.0} & 62.4~\gbf{+2.8}  & 73.0~\rbf{-1.6} \\ 
MoCo~\cite{moco}      & 32.3~\rbf{-0.6}   & 59.3~\rbf{-0.3}    & 75.3~\gbf{+0.7} \\ 
MoCov2~\cite{mocov2}      & 33.9~\gbf{+1.0} & 60.8~\gbf{+1.2}  & 75.7~\gbf{+1.1} \\ \hline
DetCo        & \bf{34.7}~\gbf{+1.8} & \bf{63.2}~\gbf{+3.6}  & \bf{76.5}~\gbf{+1.9} \\ 
\end{tabular}}
\caption{\textbf{DetCo \textit{vs.} supervised and other unsupervised methods on Cityscapes dataset.} 
All methods are pretrained 200 epochs on ImageNet.
We evaluate instance segmentation and semantic segmentation tasks.}\label{tab:city}
\vspace{-3mm}
\end{table}

\begin{table*}[]
\centering
\scalebox{0.75}{
\begin{tabular}{l|l l l |l l l |l l l |l l l}
\multirow{2}{*}{Method} &\multicolumn{3}{c|}{RetinaNet R50 12k} &\multicolumn{3}{c|}{RetinaNet R50 90k} &\multicolumn{3}{c|}{RetinaNet R50 180k} &\multicolumn{3}{c}{Keypoint RCNN R50 180k} \\ \cline{2-13} 
 &AP &AP$_{50}$ &AP$_{75}$ &AP &AP$_{50}$ &AP$_{75}$  &AP &AP$_{50}$ &AP$_{75}$ &AP$^{kp}$ &AP$^{kp}_{50}$ &AP$^{kp}_{75}$ \\ \shline
\gray{Rand Init} &4.0            &7.9            &3.5            &24.5           &39.0           &25.7           &32.2            &49.4            &34.2            &65.9           &86.5           &71.7 \\ 
Supervised  &24.3           &40.7           &25.1           &37.4           &56.5           &39.7           &38.9            &58.5            &41.5            &65.8           &86.9           &71.9 \\ \hline
InsDis\cite{wu2018unsupervised} &19.0\rbf{-5.3} &32.0\rbf{-8.7} &19.6\rbf{-5.5} &35.5\rbf{-1.9} &54.1\rbf{-2.4} &38.2\rbf{-1.5} &38.0\rbf{-0.9}  &57.4\rbf{-1.1}  &40.5\rbf{-1.0}  &66.5\gbf{+0.7} &87.1\gbf{+0.2} &72.6\gbf{+0.7} \\ 
PIRL\cite{pirl}        &19.0\rbf{-5.3} &31.7\rbf{-9.0} &19.8\rbf{-5.3} &35.7\rbf{-1.7} &54.2\rbf{-2.3} &38.4\rbf{-1.3} &38.5\rbf{-0.4}  &57.6\rbf{-0.9}  &41.2\rbf{-0.3}  &66.5\gbf{+0.7} &87.5\gbf{+0.6} &72.1\gbf{+0.2} \\ 
SwAV\cite{swav}        &19.7\rbf{-4.6} &34.7\rbf{-6.0} &19.5\rbf{-5.6} &35.2\rbf{-2.2} &54.9\rbf{-1.6} &37.5\rbf{-2.2} &38.6\rbf{-0.3}  &58.8\gbf{+0.3}  &41.1\rbf{-0.4}  &66.0\gbf{+0.2} &86.9\rbf{0.0}  &71.5\rbf{-0.4} \\ 
MoCo\cite{moco}     &20.2\rbf{-4.1} &33.9\rbf{-6.8} &20.8\rbf{-4.3} &36.3\rbf{-1.1} &55.0\rbf{-1.5} &39.0\rbf{-0.7} &38.7\rbf{-0.2}  &57.9\rbf{-0.6}  &41.5\rbf{0.0}   &66.8\gbf{+1.0} &87.4\gbf{+0.5} &72.5\gbf{+0.6} \\ 
MoCov2\cite{mocov2}     &22.2\rbf{-2.1} &36.9\rbf{-3.8} &23.0\rbf{-2.1} &37.2\rbf{-0.2} &56.2\rbf{-0.3} &39.6\rbf{-0.1} &39.3\gbf{+0.4}  &58.9\gbf{+0.4}  &42.1\gbf{+0.6}  &66.8\gbf{+1.0} &87.3\gbf{+0.4} &73.1\gbf{+1.2} \\ \hline
DetCo          &\bf{25.3}\gbf{+1.0} &\bf{41.6}\gbf{+0.9} &\bf{26.5}\gbf{+1.4} &\bf{38.4}\gbf{+1.0} &\bf{57.8}\gbf{+1.3} &\bf{41.2}\gbf{+1.5} &\bf{39.7}\gbf{+0.8}  &\bf{59.3}\gbf{+0.8}  &\bf{42.6}\gbf{+1.1}  &\bf{67.2}\gbf{+1.4} &\bf{87.5}\gbf{+0.6} &\bf{73.4}\gbf{+1.5} \\ 

\end{tabular}}
\caption{\small{
\textbf{One-stage object detection and keypoint detection fine-tuned on COCO}. 
All methods are pretrained 200 epochs on ImageNet.
DetCo outperforms all supervised and unsupervised counterparts. 
}
}
\label{tab:retina_pose}
\end{table*}

\begin{table*}[]
\centering
\scalebox{0.75}{
\begin{tabular}{l|lll|lll|lll|lll}
\multirow{2}{*}{Method} &\multicolumn{3}{c|}{RetinaNet R50 COCO 1\% Data} &\multicolumn{3}{c|}{RetinaNet R50 COCO 2\% Data} &\multicolumn{3}{c|}{RetinaNet R50 COCO 5\% Data} &\multicolumn{3}{c}{RetinaNet R50 COCO 10\% Data} \\ \cline{2-13} 
 &AP &AP$_{50}$ &AP$_{75}$ &AP &AP$_{50}$ &AP$_{75}$  &AP &AP$_{50}$ &AP$_{75}$ &AP &AP${50}$ &AP${75}$ \\ \shline
\gray{Rand Init}     &1.4  &3.5  &1.0 &2.5  &5.6  &2.0  &3.6  &7.4   &3.0   &3.7  &7.5  &3.2 \\ 
Supervised           &8.2  &16.2 &7.2 &11.2 &21.7 &10.3 &16.5 &30.3  &15.9  &19.6 &34.5 &19.7 \\ \hline
MoCo\cite{moco}                 &7.0\rbf{-1.2} &13.5\rbf{-2.7} &6.5\rbf{-0.7} &10.3\rbf{-0.9} &19.2\rbf{-2.5} &9.7\rbf{-0.6} &15.0\rbf{-1.5} &27.0\rbf{-3.3} &14.9\rbf{-1.0} &18.2\rbf{-1.4} &31.6\rbf{-2.9} &18.4\rbf{-1.3} \\
MoCo v2\cite{mocov2}              &8.4\gbf{+0.2} &15.8\rbf{-0.4} &8.0\gbf{+0.8} &12.0\gbf{+0.8} &21.8\gbf{+0.1} &11.5\gbf{+1.2} &16.8\gbf{+0.3} &29.6\rbf{-0.7} &16.8\gbf{+0.9} &20.0\gbf{+0.4} &34.3\rbf{-0.2} &20.2\gbf{+0.5} \\ \hline
DetCo            &\textbf{9.9}\gbf{+1.7} &\textbf{19.3}\gbf{+3.1} &\textbf{9.1}\gbf{+1.9} &\textbf{13.5}\gbf{+2.3} &\textbf{25.1}\gbf{+3.4} &\textbf{12.7}\gbf{+2.4} &\textbf{18.7}\gbf{+2.2} &\textbf{32.9}\gbf{+2.6} &\textbf{18.7}\gbf{+2.8} &\textbf{21.9}\gbf{+2.3} &\textbf{37.6}\gbf{+3.1} &\textbf{22.3}\gbf{+2.6}

\end{tabular}}
\caption{\small{
\textbf{Semi-Supervised one-stage detection fine-tuned on COCO 1\%, 2\%, 5\% and 10\%  data}. All methods are pretrained 200 epochs on ImageNet. 
DetCo is significant better than supervised~/~unsupervised counterparts in all metrics.
}
}
\label{tab:retina_1_2_5_10}
\end{table*}

\begin{table}[!t]
\centering
\scalebox{0.9}{
\begin{tabular}{l|c|l l l}
Method & Epoch & AP & AP$_{50}$ & AP$_{75}$ \\ \shline
\gray{Rand Init} & - &33.8 &60.2 &33.1 \\ 
Supervised                       & 90  &53.5           &81.3           &58.8 \\ \hline
InsDis~\cite{wu2018unsupervised} & 200 &55.2\gbf{+1.7} &80.9\rbf{-0.4} &61.2\gbf{+2.4} \\ 
PIRL~\cite{pirl}                 & 200 &55.5\gbf{+2.0} &81.0\rbf{-0.3} &61.3\gbf{+2.5} \\ 
SwAV~\cite{swav}                 & 800 &56.1\gbf{+2.6} &82.6\gbf{+1.3} &62.7\gbf{+3.9} \\ 
MoCo~\cite{moco}                 & 200 &55.9\gbf{+2.4} &81.5\gbf{+0.2} &62.6\gbf{+3.8} \\ 
MoCov2~\cite{mocov2}             & 200 &57.0\gbf{+3.5} &82.4\gbf{+1.1} &63.6\gbf{+4.8} \\ 
MoCov2~\cite{mocov2}             & 800 &57.4\gbf{+3.9} &82.5\gbf{+1.2} &64.0\gbf{+5.2} \\ \hline
DetCo                            & 100 &57.4\gbf{+3.9} &82.5\gbf{+1.2} &63.9\gbf{+5.1} \\ 
                            & 200 &57.8\gbf{+4.3} &82.6\gbf{+1.3} &64.2\gbf{+5.4} \\ 
                            & 800 &\bf{58.2}\gbf{+4.7} &\bf{82.7}\gbf{+1.4} &\bf{65.0}\gbf{+6.2} \\ 
\end{tabular}}
\caption{\textbf{Object Detection finetuned on PASCAL VOC07+12 using Faster RCNN-C4.} DetCo-100ep is on par with previous state-of-the-art, and DetCo-800ep achieves the best performance.
}
\label{tab:voc}
\end{table}

\vspace{2mm}
\noindent \textbf{Discussions.}
We compared the performance when transferred to object detection at different dataset scales and finetuning iterations. 
First, DetCo largely boosts up the performance of the supervised method on small datasets (\eg PASCAL VOC). 
Second, DetCo also has large advantages with COCO 12k iterations. It indicates that DetCo can fasten training converge compared with other unsupervised and supervised methods. 
Third, even with enough data (\eg COCO), Detco still significantly improves the performance compared to other unsupervised and supervised counterparts.
Finally, DetCo is friendly for detection tasks while it does not sacrifice the classification compared with concurrent SSL methods.

\subsection{Segmentation and Pose Estimation}

\noindent \textbf{Multi-Person Pose Estimation.}
The last column of Table~\ref{tab:retina_pose} shows the results of COCO keypoint detection results using Mask RCNN. DetCo also surpasses other methods in all metrics, \textit{e.g.} \textbf{1.4} AP$^{kp}$ and 1.5 AP$^{kp}_{75}$ higher than supervised counterpart.

\noindent \textbf{Segmentation on Cityscapes.}
Cityscapes is a dataset for autonomous driving in the urban street. We follow MoCo to evaluate on instance segmentation with Mask RCNN and semantic segmentation with FCN-16s~\cite{fcn}.
The results are shown in Table~\ref{tab:city}.

Although its domain is totally different from COCO, DetCo pre-training can still significantly improve the transfer performance.
On instance segmentation, DetCo outperforms supervised counterpart and MoCo v2 by \textbf{3.6} and \textbf{2.4} on AP$_{50}^{mk}$. 
On semantic segmentation, DetCo is also 1.9\% and 0.8\% higher than supervised method and MoCo v2.

\vspace{2mm}
\noindent \textbf{DensePose.}
Estimating 3D shape from a single 2D image is challenging. It can serve as a good testbed for self-supervised learning methods, so we evaluate DetCo on COCO DensePose~\cite{densepose} task and find DetCo also transfer well on this task. 
As shown in Table~\ref{tab:dprcnn}, DetCo significantly outperforms ImageNet supervised method and MoCo v2 in all metrics, especially \textbf{+1.4} on AP$_{50}$.

\subsection{Image Classification}
We follow the standard settings: ImageNet linear classification and VOC SVM classification. For ImageNet linear classification, the training epoch is 100, and the learning rate is 30, the same as MoCo.
 Our DetCo also outperforms its strong baseline MoCo v2 by \textbf{+1.1\%} in Top-1 Accuracy as shown in Table~\ref{tab:results_imgnet}. It is also competitive on VOC SVM classification accuracy compared with state-of-the-art counterparts.

\begin{table}[t]
\centering
\scalebox{0.85}{
\begin{tabular}{l|c|cc|c}
\multirow{2}{*}{Method} & \multirow{2}{*}{Epoch} & \multicolumn{2}{c|}{ImageNet} & VOC07 \\ \cline{3-5} 
 & & \multicolumn{1}{c}{Top1} & Top5 & \multicolumn{1}{c}{Acc} \\ \shline
Jigsaw~\cite{noroozi2016unsupervised} & - & 44.6 & - &64.5 \\
Rotation~\cite{gidaris2018unsupervised} & - & 55.4 &- & 63.9 \\
InsDis~\cite{wu2018unsupervised} & 200 & 56.5 & - &76.6 \\
LocalAgg~\cite{localagg} & 200 & 58.8 & - &- \\
PIRL~\cite{pirl} & 800 & 63.6 &- &81.1 \\
SimCLR~\cite{simclr} & 1000 & 69.3 & 89.0 &- \\
BYOL~\cite{byol} & 1000 & 74.3 & 91.6 & -\\
SwAV~\cite{swav} & 200 & 72.7 & - & 87.6\\
MoCo~\cite{moco} & 200 & 60.6 & - &79.2 \\ 
MoCov2~\cite{mocov2} & 200 & 67.5 & -  & 84.1\\ \hline
DetCo & 200 & 68.6 & 88.5 &85.1 \\
\end{tabular}}
\caption{\textbf{Comparison  of ImageNet Linear Classification and VOC SVM Classification.} Although DetCo is designed for detection, it is also robust and competitive on classification task, and it substantially exceeds MoCov2 baseline by 1.1\%.
} 
\label{tab:results_imgnet}
\end{table}

\begin{figure}[]
\begin{center}
\includegraphics[width=0.45\textwidth]{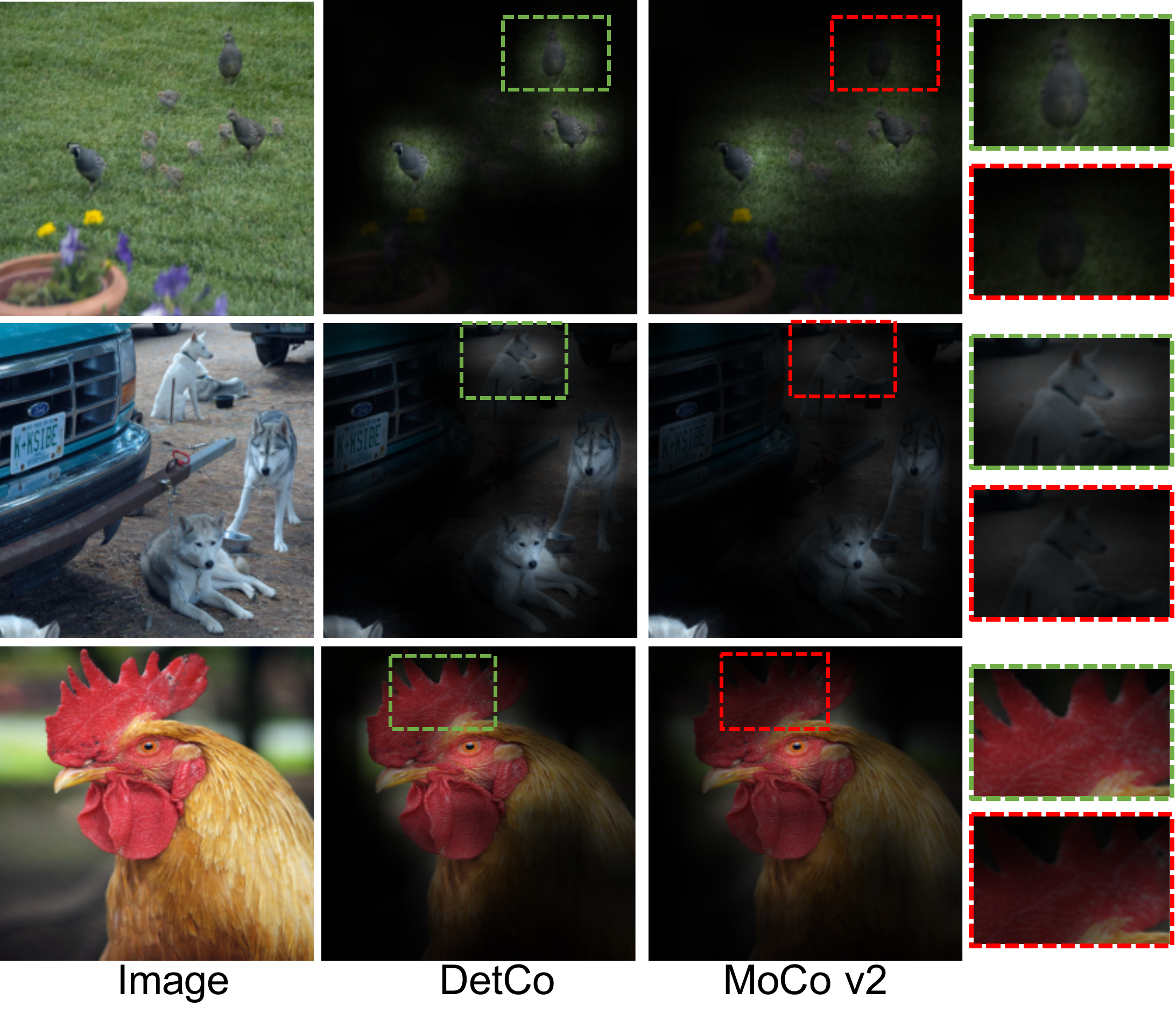}
\caption{\textbf{Attention maps generated by DetCo and MoCov2~\cite{mocov2}.} DetCo can activate more accurate object regions in the heatmap than MoCov2. More visualization results are in Appendix.}
\label{fig:heatmap}
\end{center}
\end{figure}

\noindent \textbf{Discussion.} 
While DetCo is designed for object Detection,  its classification accuracy is still competitive. On ImageNet classification, DetCo largely outperforms concurrent DenseCL~\cite{densecl}, PatchReID~\cite{dupr} and InstLoc~\cite{instloc}, even surpasses the MoCo v2 baseline~\cite{mocov2} by 1.1\%. Although inferior to strongest classification method, SwAV, DetCo exhibits better detection accuracy. Overall, DetCo achieves best classification-detection trade-off.

\subsection{Visualization Results}
Figure~\ref{fig:heatmap} visualizes the attention map of DetCo and MoCo v2. 
We can see when there is more than one object in the image, DetCo successfully locates all the objects, while MoCo v2 fails to activate some objects. 
Moreover, in the last column, the attention map of DetCo is more accurate than MoCo v2 on the boundary. 
It reflects from the side that the localization capability of DetCo is stronger than MoCo v2, which is beneficial for object detection. More analysis, implementation details and visualization results are shown in Appendix.

\subsection{Ablation Study}\label{sec:ab}
\noindent \textbf{Experiment Settings.}
We conduct all the controlled experiments by training 100 epochs. We adopt MoCo v2 as our strong baseline. More ablation studies about hyper-parameters are shown in Appendix. 
In table~\ref{tab:ab1} and~\ref{tab:ab2}, \textit{``MLS''} means \textbf{M}ulti-\textbf{L}evel \textbf{S}upervision, and \textit{``GLC''} means \textbf{G}lobal and \textbf{L}ocal \textbf{C}ontrastive learning.

\vspace{2mm}
\noindent \textbf{Effectiveness of Multi-level Supervision}.
As shown in Table~\ref{tab:ab1}~(a) and (b), when only adding the multi-level supervision on MoCo v2, the classification accuracy \textit{drop} but detection performance \textit{increase}. 
This is reasonable and expectable because for image classification, it is not necessary for each layer to remain discriminative, and only the final layer feature should be discriminative. 
However, keeping multiple level features discriminative is essential for object detection because modern detectors predict boxes in feature pyramids.
We find that intermediate supervision will slightly decrease the final layer feature's representation and improve the shallow layer features' representation, which is beneficial to object detection.

We also evaluate the VOC SVM classification accuracy at four stages: $\tt{Res2, Res3, Res4, Res5}$ to demonstrate the enhancement of the intermediate feature. As shown in Table~\ref{tab:ab2}~(a) and (b), the discrimination ability of shallow features vastly improves compared with baseline. 

\vspace{2mm}
\noindent\textbf{Effectiveness of Global and Local Contrastive Learning.}
As shown in Table~\ref{tab:ab1}~(a) and (c), when only adding global and local contrastive learning, the performance of both classification and detection boosts up and surpasses MoCo v2 baseline.  
Moreover, as shown in Table~\ref{tab:ab1}~(d), GLC can further improve the detection accuracy as well as the classification accuracy. 
This improvement mainly benefits from the GLC successfully make network learn the image-level and patch-level representation, which is beneficial for object detection and image classification.
From  table~\ref{tab:ab2}~(a),~(c) and (d), the GLC can also improve the discrimination of different stages.

\begin{table}[]
\centering
\scalebox{0.95}{
\begin{tabular}{c|c|c|l l|l}
 & +MLS & +GLC & Top1 & Top5 & mAP \\ \shline
(a) & $\times$ & $\times$ & 64.3 & 85.6 & 56.3 \\ \shline
(b) & \checkmark & $\times$ & 63.2~$\red{\pmb{\downarrow}}$ & 84.9~$\red{\pmb{\downarrow}}$ & 57.0~$\green{\pmb{\uparrow}}$ \\ \hline
(c) & $\times$ &  \checkmark & 67.1~$\green{\pmb{\uparrow}}$ & 87.5~$\green{\pmb{\uparrow}}$ & 56.8~$\green{\pmb{\uparrow}}$ \\ \hline
(d) & \checkmark & \checkmark & 66.6~$\green{\pmb{\uparrow}}$ & 87.2~$\green{\pmb{\uparrow}}$ & 57.4~$\green{\pmb{\uparrow}}$ \\ 
\end{tabular}}
\caption{\textbf{Ablation:~multi-level supervision~(MLS) and global and local contrastive learning~(GLC)}. The results are evaluated on ImageNet linear classification and PASCAL VOC07+12 detection.}
\label{tab:ab1}
\end{table}
\begin{table}[]
\centering
\scalebox{0.9}{
\begin{tabular}{c|c|c|l l l l}

    & +MLS & +GLC & Res2 & Res3 & Res4 & Res5 \\ \shline
(a) & $\times$    & $\times$    & 47.1 & 58.2 & 70.9 & 82.1 \\ \shline
(b) & \checkmark    & $\times$    & 50.9~$\green{\pmb{\uparrow}}$ & 67.1~$\green{\pmb{\uparrow}}$ & 78.7~$\green{\pmb{\uparrow}}$ & 81.8~$\red{\pmb{\downarrow}}$ \\ \hline
(c) & $\times$  & \checkmark    & 47.8~$\green{\pmb{\uparrow}}$ & 59.8~$\green{\pmb{\uparrow}}$ & 75.0~$\green{\pmb{\uparrow}}$ & 84.6~$\green{\pmb{\uparrow}}$ \\ \hline
(d) & \checkmark    & \checkmark    & 51.6~$\green{\pmb{\uparrow}}$ & 69.7~$\green{\pmb{\uparrow}}$ & 82.5~$\green{\pmb{\uparrow}}$ & 84.3~$\green{\pmb{\uparrow}}$ \\ 
\end{tabular}
}
\caption{\textbf{Ablation:~multi-level supervision~(MLS) and global and local contrastive learning~(GLC)}. Accuracy of feature in different stages are evaluated by PASCAL VOC07 SVM classification.
}
\label{tab:ab2}
\end{table}

\section{Conclusion and Future work}
This work presents DetCo, a simple yet effective pretext task that can utilize large-scale unlabeled data to provide a pre-train model for various downstream tasks.
DetCo inherits the advantage of strong MoCo v2 baseline and beyond it by adding (1) multi-level supervision (2) global and local contrastive learning.
It demonstrates state-of-the-art transfer performance on various instance-level detection tasks, \eg VOC and COCO detection as well as semantic segmentation, while maintaining the competitive performance on image classification. 
We hope DetCo can serve as an alternative and useful pre-train model for dense predictions and faciltate future research. 

\noindent \textbf{Acknowledgement}.
We would like to thank Huawei to support $>$200 GPUs and Yaojun Liu, Ding Liang for insightful discussion without which this paper would not be possible.

{\small
\bibliographystyle{ieee_fullname}
\bibliography{egbib}
}

\end{document}